\documentclass[10pt,twocolumn,letterpaper]{article}

\usepackage{iccv}
\usepackage{times}
\usepackage{epsfig}
\usepackage{graphicx}
\usepackage{amsmath}
\usepackage{amssymb}

\usepackage{multirow}
\usepackage{adjustbox}
\usepackage{tabu}
\usepackage{xcolor}
\usepackage{soul}
\usepackage[linesnumbered, ruled]{algorithm2e}
\usepackage{algpseudocode}%
\usepackage{bm}
\usepackage{array}
\usepackage{enumitem}
\usepackage{verbatim}
\usepackage{subfigure}
\usepackage{caption}
\usepackage{algorithmicx}
\usepackage{comment}
\usepackage{lipsum}
\usepackage{stringstrings}

\usepackage[margin=4pt,font=footnotesize,labelfont=bf,labelsep=endash,tableposition=top]{caption}

\usepackage[pagebackref=true,breaklinks=true,letterpaper=true,colorlinks,bookmarks=false]{hyperref}

\iccvfinalcopy %

\def\iccvPaperID{2919} %

\ificcvfinal\fi

\begin{document}

\title{Self-training with progressive augmentation  for  unsupervised cross-domain person re-identification\thanks{Work was done when X. Zhang  was visiting The University of Adelaide. 
First two authors contributed to this work equally. 
C. Shen is the corresponding author: $\sf chunhua.shen@adelaide.edu.au $
}
}

\author{
Xinyu Zhang$^{1}$ \quad  ~
Jiewei Cao$^{2}$ \quad   ~
Chunhua Shen$^{2}$ \quad ~
Mingyu You$^{1}$ \quad \\
$^{1}$Tongji University, China 
~
~
~
$^{2}$The University of Adelaide, Australia\\
}

\maketitle
\ificcvfinal\thispagestyle{empty}\fi

\begin{abstract}
Person re-identification (Re-ID) has achieved %
great 
improvement with deep learning and a large amount of labelled training data. 
However, it remains a  challenging task %
for 
adapting
a model trained %
in
a source domain of labelled data to a  target domain of only unlabelled data available. 
In this work, we develop a self-training method with progressive augmentation framework (PAST) to promote the model performance progressively on the target dataset. 
Specially, our PAST framework consists of two stages, namely, conservative stage and promoting stage. The conservative stage captures the local structure of target-domain data points with triplet-based loss functions, leading to improved feature representations. 
The promoting stage continuously optimizes the network 
by appending a changeable classification layer to the last layer of the model,
enabling the use of  
global information about the data distribution. 
Importantly, we propose a new self-training strategy that progressively augments the model capability
by adopting conservative and promoting stages alternately. 
Furthermore, to improve the reliability of selected triplet samples, we introduce a ranking-based triplet loss in the conservative stage, which is a label-free objective function basing on the similarities between data pairs.
Experiments demonstrate that the proposed method achieves %
state-of-the-art person Re-ID performance under the unsupervised cross-domain setting.

Code is available at: \href{https://tinyurl.com/PASTReID}{{ $\sf tinyurl.com/PASTReID$}}
\end{abstract}

%
\section{Introduction}
Person re-identification (Re-ID) is a crucial task in surveillance and security, which aims to locate a target pedestrian across non-overlapping camera views using a probe image.
With the advantages of convolutional neural networks (CNN), many person Re-ID works focus on \emph{supervised learning}~\cite{DTL, PCB, AACN, mask2stream, CamStyle, MLFN, MTDnet, HA-CNN, svdnet, DPFL,paisitkriangkrai2015learning} and achieve satisfactory improvements. Despite the great success, they depend on large labelled datasets which are costly and
sometime impossible
to obtain.

\begin{figure}[t!]
\centering
\includegraphics[trim =0mm 0mm 0mm 0mm, clip, width=.67951\linewidth]{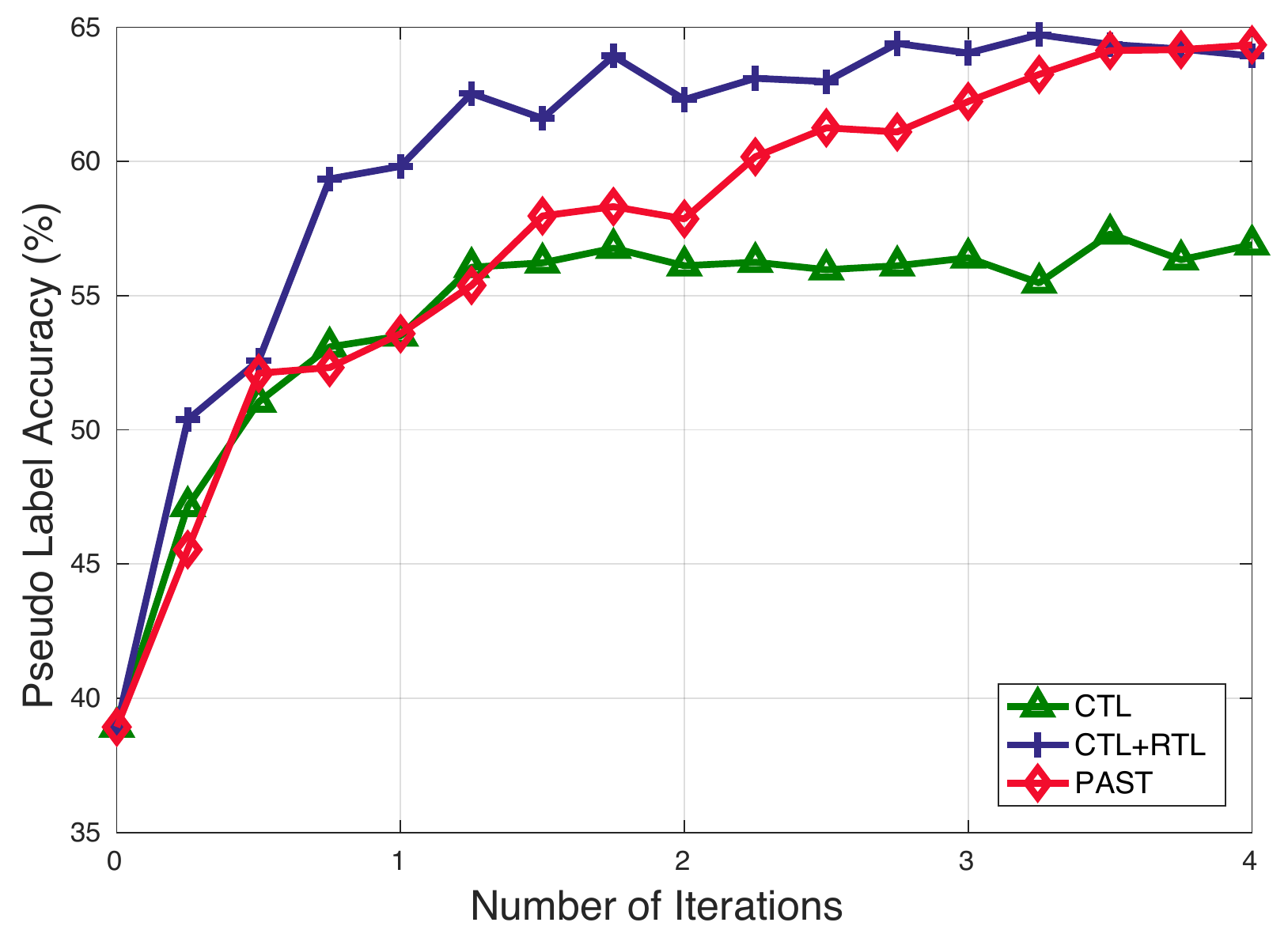}
\includegraphics[trim =0mm 0mm 0mm 0mm, clip, width=.67951\linewidth]{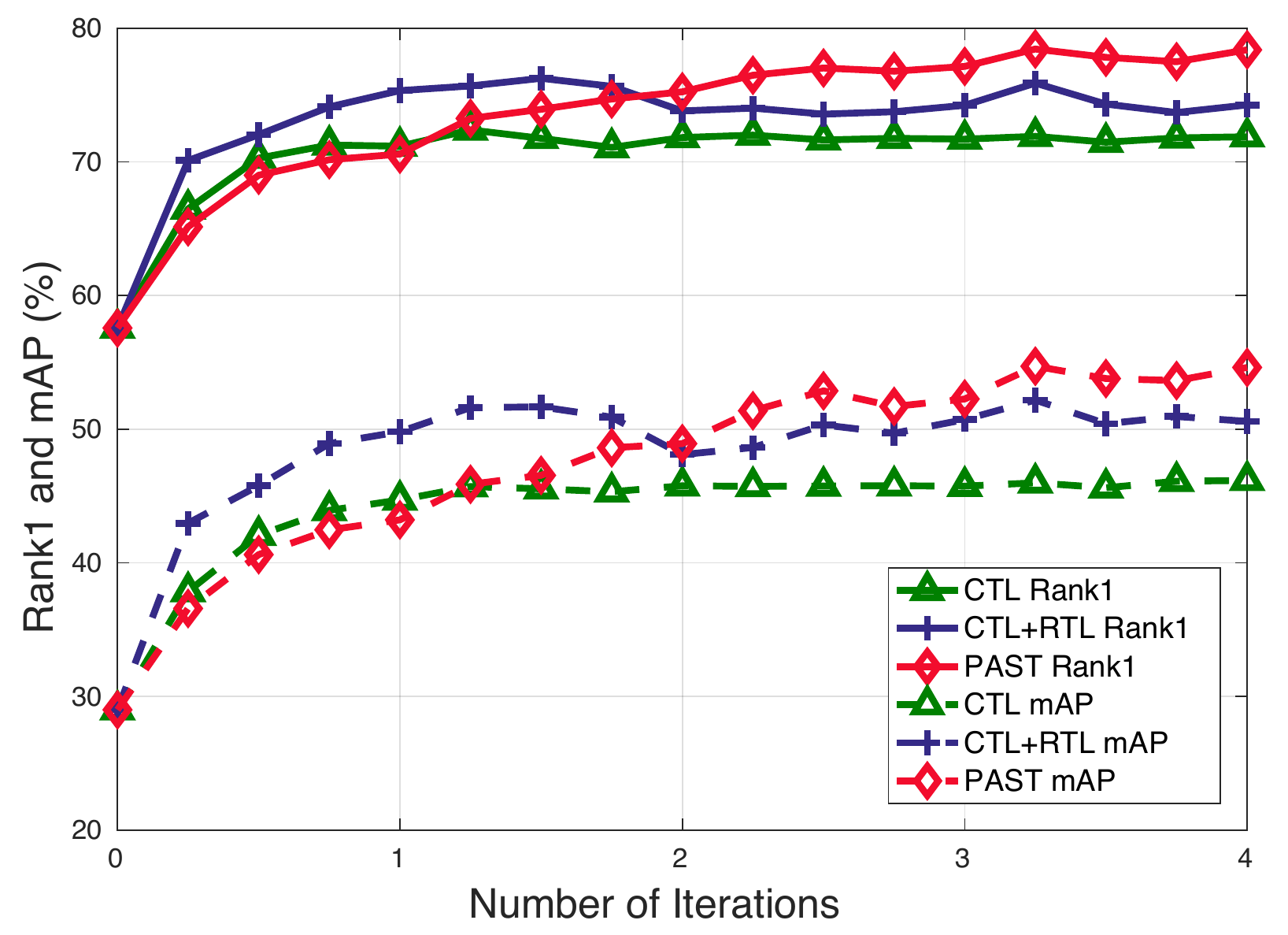}
\setlength{\abovecaptionskip}{0.1cm} 
\caption{Label quality \vs\ model generalization. The accuracy of pseudo labels prediction (top) and performance comparison (bottom) of different training processes over training iterations. Here we use Duke~\cite{duke} as the source domain and Market-1501~\cite{market1501} as the target domain.}
\label{fig:figure1}
\vspace{-5mm}
\end{figure}

To %
tackle this problem, 
a few \emph{unsupervised learning} methods~\cite{TJ-AIDL, stepwise, BUC} propose to take advantage of abundant unlabelled data, which are easier to collect in general. Unfortunately, due to lack of supervision information, the performance of unsupervised methods is typically weak, thus being less effective for practical usages.
In contrast, \emph{unsupervised cross-domain} methods~\cite{PTGAN, SPGAN, TJ-AIDL, HHL, EANet, UMDL, PUL, tfusion, ARN, theory} propose to use both labelled datasets (source domain) and unlabelled datasets (target domain).
However, directly applying the models trained in the source domain to the target domain leads to unsatisfactory performances due to the inconsistent characteristics between the two domains, which is known as the \emph{domain shift} problem~\cite{ARN}.
In unsupervised cross-domain Re-ID, the problem becomes how to transfer the learned information of a pre-trained model from the source domain to the target domain effectively in an unsupervised manner.

Some domain transfer methods~\cite{HHL, EANet, UMDL, PUL, tfusion, ARN, theory, stepwise} have taken great efforts to address this challenge, where the majority are based on \emph{pseudo label} estimation~\cite{PUL, theory, tfusion}.
They extract embedding features of unlabelled target datasets from the pre-trained model and apply unsupervised clustering methods (\eg, $k$-means and $\text{DBSCAN}$~\cite{DBSCAN}) to separate the data into different clusters. 
The samples in the same cluster are assumed to belong to the same person, which are adapted for training as in supervised learning.
The drawback of these methods is that the performance highly depends on the clustering quality, reflecting on whether samples with the same identity are assigned to one cluster. 
In other words, performance relies on to what extent are the pseudo labels from clustering consistent with ground truth identity labels.
Since the percentage of corrupted labels largely affect the model generalization on the target dataset~\cite{generalization}, we propose a method to improve the quality of labels in a progressive way which results in considerable improvement of model generalization on the unseen target dataset.

Here 
we propose a 
new 
\emph{Self-Training with Progressive Augmentation framework} (PAST) to: 1) restrain error amplification at early training epochs when the quality of pseudo label can be low; and 2) progressively incorporate more confidently labelled examples for self-training when the label quality is becoming better. PAST has two learning stages, \ie, \emph{conservative} and \emph{promoting} stage, which consider complementary data information via different learning strategies for self-training.

\noindent\textbf{Conservative Stage.}
As shown in Figure~\ref{fig:figure1}, the percentage of correctly labelled data is low at first due to the domain shift. In this scenario, we need to select \textit{confidently labelled} examples to reduce label noise. We consider the similarity score between images as a good indicator of confidence measure. Beside 
the widely used clustering-based triplet loss (CTL) ~\cite{batchhardtriplet}, which is sensitive to the quality of pseudo labels generated from clustering method, we propose a novel label-free loss function, \emph{ranking-based triplet loss (RTL)}, to better capture the characteristic of data distribution in the target domain.

Specifically, we calculate the ranking score matrix for the whole target dataset and generate triplets by selecting the positive and negative examples from the top $\eta$ and  $\left(\eta, 2\eta\right]$ ranked images for each anchor. The triplets are then fed into the model and trained with the proposed RTL. 
In the conservative stage, we mainly consider the local structure of data distribution which is crucial for avoiding model collapse when the label quality is mediocre at early learning epochs.

\noindent\textbf{Promoting Stage.}
However, as the number of training triplets dramatically grows in large datasets and triplets only focus on local information, the learning process with triplet loss inevitably becomes instability and suffers from the local-optimal result, as shown by the ``CTL'' and ``CTL+RTL'' in Figure~\ref{fig:figure1}.
To remedy this issue, we propose to use the global distribution of data points for network training at the promoting stage.
That is, we treat each cluster as a class and convert the learning process into a classification problem. Softmax cross-entropy loss is used to force different categories staying apart for encouraging inter-class separability. After the promoting stage, the model is prone to be more stable which facilitates learning the discriminative features.
Since the error is most likely amplified when training on images with extremely corrupted labels using the softmax cross-entropy loss, we employ this stage following the conservative learning stage and carry out two stages interchangeably. With this alternate process, our proposed PAST framework can stabilize the training process and progressively improve the capability of model generalization on the target domain. 

To summarize, our main contributions are as follows.
\setlist{nolistsep}
\begin{itemize}[fullwidth, itemindent=1em]
\item[1)] We present a novel self-training with progressive augmentation framework (PAST) to solve the unsupervised cross-domain person Re-ID problem. By executing the two-stage self-training process, namely,
conducting conservative and promoting stage alternately, our method considerably improve the model generalization on unlabelled target-domain datasets.
\item[2)] We propose a ranking-based triplet loss (RTL), solely relying on similarity scores of data points, to avoid selecting triplet samples using unreliable pseudo labels.
\item[3)] We take advantage of global data distribution for model training with softmax cross-entropy loss, which is  beneficial for training stability and promoting the capability of model generalization.
\item[4)]  %
Experimental results on three large-scale datasets indicate the effectiveness of our proposed method on the task of unsupervised cross-domain person Re-ID.
\end{itemize}
\setlist{nolistsep}

\begin{figure*}[t]
\centering
\includegraphics[trim =0mm 0mm 0mm 0mm, clip, width=.9851\linewidth]{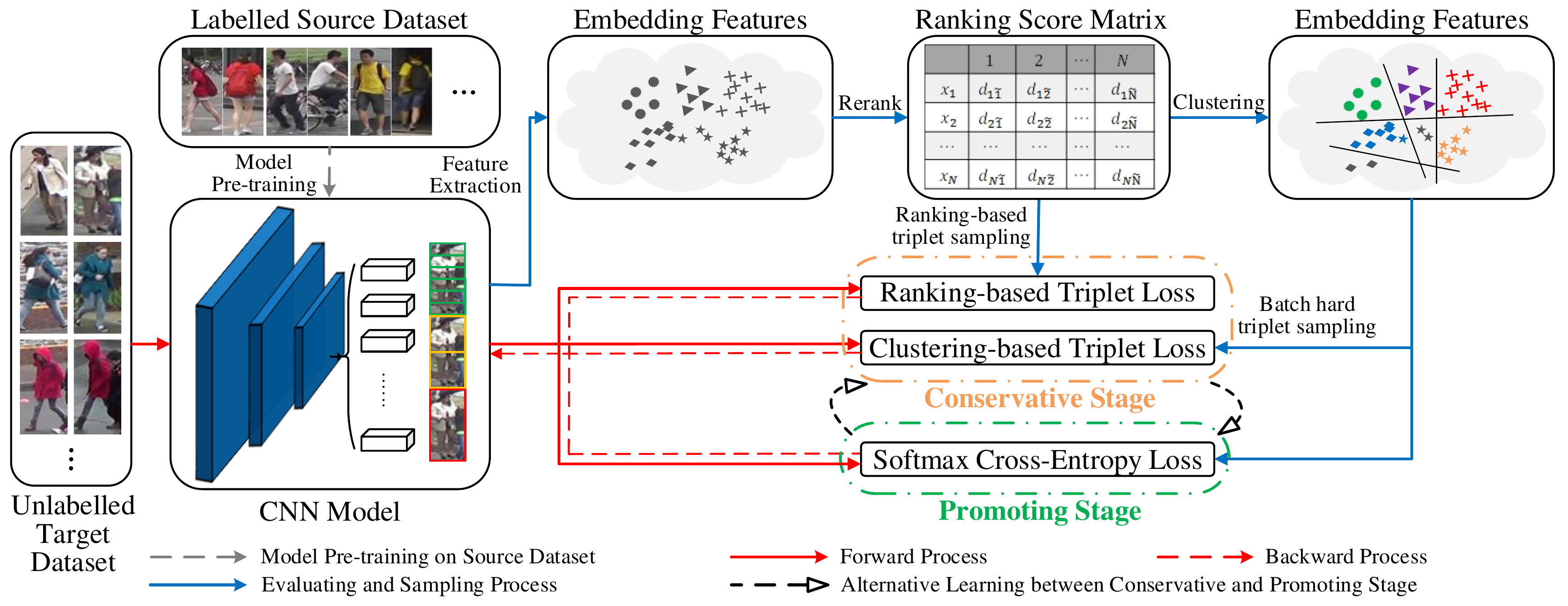}
\setlength{\abovecaptionskip}{0.013cm} 
\caption{The overview of our self-training framework with progressive augmentation (PAST). The model is pre-trained on the labelled source dataset. During training, we first carry out a sampling process, which consists of extracting embedding features of unlabelled target dataset with the current model and calculating the ranking score matrix with Eq.~\eqref{eq:similarity}. We then assign pseudo labels to training samples via HDBSCAN~\cite{HDBSCAN} clustering method. 
After that, we conduct conservative stage by using clustering-based triplet loss (CTL) and the proposed ranking-based triplet loss (RTL) simultaneously to update the model. 
In promoting stage, the softmax cross-entropy loss is employed to further improve the capability of the model. 
Note that the conservative stage and promoting stage alternate iteratively during the whole learning process.
For Re-ID evaluation, we extract the embedding features for both query and gallery images and use the cosine distance for ranking.} 
\label{fig:model}
\end{figure*}

\subsection{Related Work}
\noindent\textbf{Supervised Person Re-ID.} Most existing deep person Re-ID methods follow a supervised setting. They are mainly based on either well-designed model architectures~\cite{PCB, multichannel, cascaded, multiregion, AlignedReID, AACN, DPFL, MLFN}, additional attributions~\cite{mask2stream, MGCAM, CamStyle, MTDnet} or metric learning~\cite{DTL, batchhardtriplet, lisanti2015person, rerank}. Although significant progress has been obtained by these methods, they all require a large amount of labelled training data, which are costly and impractical to be annotated due to drastic appearance change among different datasets. 

\noindent\textbf{Unsupervised Person Re-ID.} To alleviate the above limitation, unsupervised person Re-ID methods~\cite{CAMEL, stepwise, BUC, WangGX14, WangZXG16} are proposed to make full use of large-scale unlabelled datasets. Most of them exploit cross-view identity-specific information to capture discriminate features~\cite{CAMEL, WangZXG16} or incorporate clustering methods into training to separate unlabelled images into different classes~\cite{BUC, stepwise}. However, since specific identity labels are unavailable, these unsupervised learning methods are not able to achieve  comparable results as supervised-based approaches.

\noindent\textbf{Unsupervised Cross-Domain Person Re-ID.} Recently, researchers pay intensive attention to unsupervised cross-domain person Re-ID algorithms~\cite{PTGAN, SPGAN, TJ-AIDL, HHL, EANet, UMDL, PUL, tfusion, ARN, theory} to leverage Re-ID models pre-trained in the source domain to improve the performance on unlabelled target domain. They all focus on overcoming domain shift so as to learn domain-invariant feature representation. 

Among these existing works, PTGAN~\cite{PTGAN} and SPGAN~\cite{SPGAN}  transfer source images into target-domain style by CycleGAN and then use translated images to train a model. However, due to unable to guarantee the identity of  generated images, these style transfer learning methods can not result in satisfactory performance. 
Another line of unsupervised cross-domain person Re-ID works~\cite{TJ-AIDL, HHL, tfusion, EANet} combine other auxiliary information as an assistant task to improve the model generalization. 
For instance, TFusion~\cite{tfusion} integrates spatio-temporal patterns to improve the Re-ID precision, while EANet~\cite{EANet} uses pose segmentation. 
TJ-AIDL~\cite{TJ-AIDL} learns an attribute-semantic and identity discriminative feature representation space simultaneously, which can be transferred to any new target domain for re-id tasks. 
Similar as the difficulty of supervised learning, these domain adaptation approaches suffer from the requirement of collecting attribute annotations.

Beyond the above methods, some approaches~\cite{PUL, theory, tfusion} focus on estimating pseudo identity labels on the target domain so as to learn deep models in a supervised manner. 
Usually, clustering methods are used in the feature space to generate a series of clusters which are used to update networks with an embedding loss (\eg, triplet loss~\cite{batchhardtriplet} or contrastive loss)~\cite{theory, tfusion} or classification loss (\eg, softmax cross-entropy loss)~\cite{PUL}. 
Whereas, embedding loss functions suffer from the limitation of sub-optimal results and slow convergence, while classification loss extremely depends on the quality of pseudo labels.
While the work in \cite{oneshot}  introduces a simple domain adaptation framework which also use both triplet loss and softmax cross-entropy loss jointly, it aims at solving one-shot leaning problem.

\section{Our Method}
For unsupervised cross-domain person Re-ID, the problem that  we concentrate on is how to learn robust feature representations for unlabelled target datasets using the prior knowledge from the labelled source datasets. In this section, we present our proposed \emph{self-training with progressive augmentation framework} (PAST) in detail. 

\subsection{Overview of Our Proposed Framework} 
The overall framework of our proposed \emph{self-training with progressive augmentation framework} (PAST) is described in Figure~\ref{fig:model}. 
The framework is based on a deep neural network $M$ trained on ImageNet~\cite{imagenet}, which contains two main components: \emph{conservative stage} and \emph{promoting stage}. 

We first fine-tune the model $M$ using labelled source training dataset $S$ in a supervised manner. 
Then, this pre-trained model is utilized to extract features $\textbf{F}$ on all training images in the target domain $T$, which are used as the input features of our framework. 
For the conservative stage, based on the ranking score matrix $D_{R}$ learned from the input features, we can generate a more reliable training set $T_{U}$ via the HDBSCAN~\cite{HDBSCAN} clustering method (other clustering methods can be employed here too). 
This updated training set $T_{U}$ is a subset of the whole training data $T$. Combining with two triplet-based loss functions, \ie, \emph{clustering-based triplet loss (CTL)} and the proposed \emph{ranking-based triplet loss (RTL)}, local structure of the current updated training set can be captured for model optimization. 
After that, we can use the new model to extract features $\textbf{F}_{U}$ of the current training set $T_{U}$.
Next, in the promoting stage, with the new features $\textbf{F}_{U}$ from the conservative stage, we propose to employ softmax cross-entropy loss for further optimizing the network. 
At this stage, the global distribution of the training set is considered to improve the discrimination of feature representation.
Finally, the capability of model generalization is improved gradually by training the network with the conservative stage and promoting stage alternately.

\subsection{Conservative Stage}
In the task of unsupervised domain adaptation, it is a natural goal to gather samples of the same identity together and push samples from different classes away from each other. 
Triplet loss~\cite{HHL, theory, tfusion} has been proved to be able to discover meaningful underlying local structure of data distribution by generating reliable triplets of the target data. 
Different from the supervised setting, pseudo labels are assigned to unlabelled samples, which is more difficult to construct high-quality triplets. 
Therefore, our goal is to design a learning strategy to not only generate reliable samples but also improve the model performance. 

In practice, we conduct the following procedure in the conservative stage. At the beginning, on the whole training dataset $T$: $\lbrace x_1, x_2, ..., x_N \rbrace$, we extract features $\mathbf{F}$: $\{ \mathbf{f}(x_1), \mathbf{f}(x_2), ..., \mathbf{f}(x_{N})\}$ from the current model, and adopt the $k$-reciprocal encoding~\cite{rerank}, which is a variation of the Jaccard distance between nearest neighbors sets, to generate the distance matrix $D$ as:
\begin{equation}
\setlength{\abovedisplayskip}{0.1cm}
\setlength{\belowdisplayskip}{0.1cm}
\begin{split}
D = [\textbf{D}_{J}(x_{1})~~\textbf{D}_{J}(x_{2})~~\ldots&~~\textbf{D}_{J}(x_{N})] ^T,\\
\textbf{D}_{J}(x_{i}) = [ d_{J}(x_{i}, x_{1})~~d_{J}(x_{i}, x_{2})~~&\ldots~~d_{J}(x_{i}, x_{N})],\\
\forall i\in \{1,~2,~\ldots,~&N\},
\end{split}
\label{eq:distance}
\end{equation}    
where $\textbf{D}_{J}(x_{i})$ represents the distance vector of one specific person $x_{i}$ with all training images. $d_{J}(x_{i}, x_{j})$ is the Jaccard distance between sample $x_i$ and $x_j$.

According to the fact that a smaller distance reflects more similarities between two images, we sort every distance vector $\textbf{D}_{J}(x_{i})$ from smallest value to largest value, yielding ranking score matrix $D_R$ as:
\begin{equation}
\setlength{\abovedisplayskip}{0.1cm}
\setlength{\belowdisplayskip}{0.1cm}
\begin{split}
D_R = [\textbf{D}_{R}(x_{1})~~\textbf{D}_{R}(x_{2})~~\ldots&~~\textbf{D}_{R}(x_{N})]^T,\\
\textbf{D}_{R}(x_{i}) = [ d_{J}(x_{i}, \widetilde{x}_{1})~~d_{J}(x_{i}, \widetilde{x}_{2})&~~\ldots~d_{J}(x_{i}, \widetilde{x}_{N})],\\
\forall i\in \{1,~2,~\ldots,~&N\},
\end{split}
\label{eq:similarity}
\end{equation}
where $\textbf{D}_{R}(x_{i})$ is the ranking format of $\textbf{D}_{J}(x_{i})$ from small to large. Given a specific sample $x_i$, $\widetilde{x}_{j}$ in $d_{J}(x_{i}, \widetilde{x}_{j})$ represents the $j$-th most similar sample.

Then, we apply a hierarchical density-based clustering algorithm (HDBSCAN)~\cite{HDBSCAN} on $D_R$ to split the whole training images into different clusters, which are considered as pseudo labels. After HDBSCAN, some images, not belonging to any clusters, are discarded. Thus, we use images with assigned labels as the updated training set $T_U$ for further model optimization. 

We combine two types of triplet loss functions together to update the model, \ie,  clustering-based triplet loss (CTL) and ranking-based triplet loss (RTL), which are different from the way of triplets selection as well as the way for model optimization. 

\textbf{Clustering-based Triplet Loss (CTL).} One loss function that we use is batch hard mining triplet loss~\cite{batchhardtriplet}, proposed to mine relations among samples within a mini-batch. We randomly sample $P$ clusters and $K$ instances in each cluster to compose a mini-batch with size of $PK$. For each anchor image $x_a$, the corresponding hardest positive sample $x_p$ and the hardest negative sample $x_n$ within the batch are selected to form a triplet. Since the pseudo labels are from a clustering method, we rename this loss function as \emph{clustering-based triplet loss} (CTL), which is formulated as,
\begin{equation}
\setlength{\abovedisplayskip}{0.1cm}
\setlength{\belowdisplayskip}{0.1cm}
\begin{aligned}
L_{CTL} = & \sum_{a=1}^{PK}[m + ||\mathbf{f}(x_a)-\mathbf{f}(x_p)||_2 - ||\mathbf{f}(x_a)-\mathbf{f}(x_n)||_2]_+\\
= &\sum_{i=1}^P\sum_{a=1}^K[m +  \overbrace{\max_{p=1...K}||\mathbf{f}(x_{i,a})-\mathbf{f}(x_{i,p})||_2}^{\rm hardest \ positive} \\ 
&- \underbrace{\min_{\substack{n=1...K\\
{j=1...P}\\
j \neq i}} ||\mathbf{f}(x_{i,a})-\mathbf{f}(x_{j,n})||_{2}}_{\rm hardest \ negative}]_+,
\end{aligned}
\label{eq:CTL}
\end{equation}
where $x_{i,j}$ is a data point representing the $j$-th image of the $i$-th cluster in the batch. $\mathbf{f}(x_{i,j})$ is the feature vector of $x_{i,j}$. 

\textbf{Ranking-based Triplet Loss (RTL).}
However, it is clear that the effect of CTL highly depends on the quality of label estimation, which is hard to decide whether the clustering result is correct or not. 
Therefore, we propose a Ranking-based Triplet Loss (RTL) to make full use of the ranking score matrix $D_R$. It is a label-free method reflecting the relation between data pairs.

Specifically, given a training anchor $x_a$, positive sample $x_p$ is randomly selected from the top $\eta$ nearest neighbors according to the ranking score vector $\textbf{D}_{R}(x_a)$, and negative sample $x_n$ is from the location $\left(\eta, 2\eta\right]$. In addition, instead of hard margin in CTL,we introduce a soft margin based on the relative ranking position of $x_p$ and $x_n$, which can adapt well to different scales of intra-class variation. The formula of RTL is shown as,
\begin{equation}
\setlength{\abovedisplayskip}{0.1cm}
\setlength{\belowdisplayskip}{0.1cm}
\begin{aligned}
L_{RTL} = & \sum_{a=1}^{PK}[\dfrac{\vert P_p - P_n\vert}{\eta}m ~+ \\
& ||\mathbf{f}(x_a)-\mathbf{f}(x_p)||_2 - ||\mathbf{f}(x_a)-\mathbf{f}(x_n)||_2]_+,
\end{aligned}
\label{eq:RTL}
\end{equation}
where the selected anchors in each batch are the same as CTL. $m$ is a basic hard margin same as Eq.~\eqref{eq:CTL}. $\eta$ is the maximum of ranking position for positive sample selection. $P_p$ and $P_n$ are the ranking positions of $x_p$ and $x_n$ with respect to $x_a$.

To summarize, we optimize the network using the combination of CTL and RTL to better capture the local-constraint information of data distribution. Our final triplet-based loss function in conservative stage is shown in Eq.~\eqref{eq:C}: 
\begin{equation}
\setlength{\abovedisplayskip}{0.1cm}
\setlength{\belowdisplayskip}{0.1cm}
\begin{aligned}
L_{C} = L_{RTL} +\lambda  L_{CTL},
\end{aligned}
\label{eq:C}
\end{equation}
where $\lambda$ is the loss weight to trade off the influence of two loss functions. Experiments show that this combined triplet-based loss function can certainly improve the capability of model representation.

\subsection{Promoting Stage}
Nevertheless, since triplet-based loss functions only focus on the data relation within each triplet, the model will be prone to instability and stuck into a suboptimal local minimum. 
To alleviate this problem, we propose to apply classification loss to further improve model generalization by taking advantage of global information of training samples. 
In the promoting stage, a fully-connected layer is added at the end of the model as a classifier layer, which is initialized according to the features of current training set. Softmax cross-entropy loss is used as the objective function, which is formulated as:
\begin{equation}
\setlength{\abovedisplayskip}{0.1cm}
\setlength{\belowdisplayskip}{0.1cm}
L_{P} = -\sum^{PK}_{i=1} \log \frac{e^{W^T_{\hat{y}_{i}}x_{i}}}{\sum_{c=1}^{C}e^{W^T_{c}x_{i}}},
\label{progressiveloss}
\end{equation}
where $\hat{y}_{i}$ is the pseudo label of the sample $x_{i}$. $C$ is the number of clusters from the  HDBSCAN clustering method with updated training set $T_{U}$.

\textbf{Feature-based Weight Initialization for Classifier.} 
Due to the variation of cluster numbers $C$, the newly added classifier layer $CL$ should be initialized every time executing HDBSCAN. Instead of random initialization, we exploit the mean features of each cluster as the initial parameters. Specifically, for each cluster $c$, we calculate the mean feature $\overline{F}_{c}$ by averaging all the embedding features of its elements. The parameters $\mathbf{W}$ of $CL$ are initialized as follows:
\begin{equation}
\setlength{\abovedisplayskip}{0.1cm}
\setlength{\belowdisplayskip}{0.1cm}
W_{c} = \overline{F}_{c}, \: c\in \{1, 2, \dots, C\},
\label{intialclassifier}
\end{equation}
where $\mathbf{W}\in \mathbb{R}^{d \times C}$, $W_{c}$ is the $c$-th column of $\mathbf{W}$, and $d$ is the feature dimensionality. 
An advantage of this initialization is that we can use the previous information to avoid the fluctuation of accuracy caused by random initialization, which is useful for the convergence of model training.

\setlength{\textfloatsep}{0.1cm}
\begin{algorithm}[t]
	\begin{footnotesize}
		\SetAlgoLined
        \SetKwInOut{Input}{Input}
        \SetKwInOut{Output}{Output}
        \SetKwInput{Initialization}{Initialization}
        \SetKwInput{Training}{Training Process}
        \Input{labelled source domain dataset $S$; 
		whole unlabelled target domain training dataset $T$; CNN model $M$ pre-trained on ImageNet; maximum iteration $I_{\max}$; HDBSCAN clustering method; minimal samples in each cluster for HDBSCAN $S_{\min}$.}
		\Output{Model $M$.}
		\Initialization{Initialize model $M$ on $S$; Initial selected training set $T_U = T$.} 
		\For{$i=1$ \KwTo $I_{\max}$}
		{
        \vspace{0.1cm}
		{\emph{Conservative Stage:}}\\
		Extract embedding features $\mathbf{F}$ on training data $T$ from $M$\;
		Compute ranking score matrix $D_R$ on whole training data $T$ with $\mathbf{F}$ according to Eq.~\eqref{eq:similarity}\;
		Update training set $T_U$ using HDBSCAN$(D_R; S_{\min})$\;
		Update model $M$ using $T_U$ according to Eq.~\eqref{eq:C}\;
		Extract embedding features $\mathbf{F}_U$ on $T_U$ from $M$\;
	   \vspace{0.1cm}
		{\emph{Promoting Stage:}}\\
		Initialize classifier layer $CL$ based on $\mathbf{F}_U$ according to Eq.~\eqref{intialclassifier}\;
		Update model $M$ with classifier layer using $T_{U}$ according to Eq.~\eqref{progressiveloss}\;
        }
		\caption{The Self-training with Progressive Augmentation Framework (PAST)}\label{algo}
	\end{footnotesize}
\end{algorithm}
\setlength{\floatsep}{0.15cm}

\subsection{Alternate  Training}
\begin{figure*}[t]
\centering
\includegraphics[trim =0mm 0mm 0mm 0mm, clip, width=.91\linewidth]{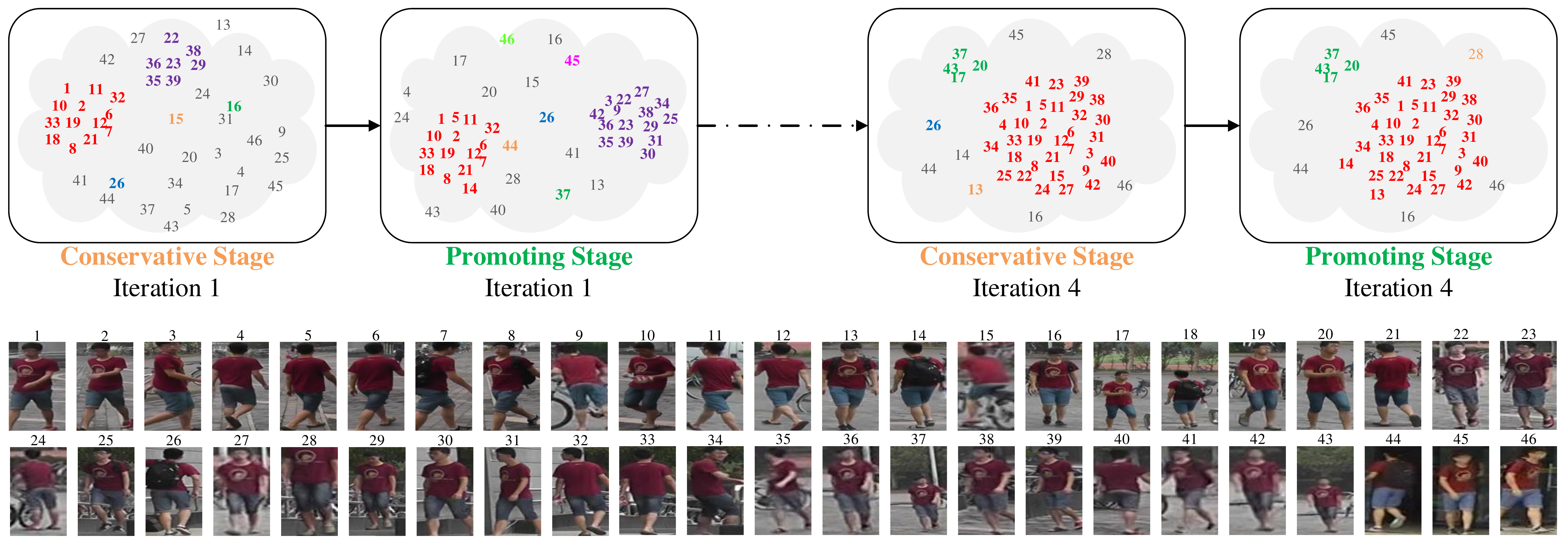}
\setlength{\abovecaptionskip}{-0.00025cm} 
\setlength{\belowcaptionskip}{-0.00055cm}
\caption{The alternate self-training process of our PAST framework on one visual example. All images belong to same person in truth. Samples with same color denotes that they are assigned to same pseudo label generated by HDBSCAN clustering method. \textcolor[rgb]{0.55,0.55,0.55}{Gray} figure means the sample not belonging to any cluster and not being used for model training. From training iteration 1 to iteration 4, more samples are selected for training. At the same time, the pseudo labels are more reliable.}
\label{fig:visual}
\end{figure*}

The learning process is expected to progressively improve the model capability of generalization, which can avoid model to fall into local optimum. In this paper, we carefully develop a simple yet effective self-training strategy which can capture local structure and global information of training images. 
That is, the conservative stage and the promoting stage are conducted alternately. 
At the beginning, the model is trained only using the local relations between data points alone, so that the difficulty of error amplification brought by softmax loss can be prevented. 
After several training steps in the conservative stage, the ability of model representation and the quality of clusters are more trusty. 
Then model capability is further augmented using Softmax cross-entropy loss in the promoting stage and the updated model is used as the initial state for conservative stage alternately. 
As the training goes on, model generalization is improved, allowing to learn more discriminate feature representation of training images. 
The details of this two-stage alternate self-training are included in Algorithm \ref{algo}. 
We also list one visual example of this alternate self-training process, shown in Figure~\ref{fig:visual}. 
It is proved that our proposed PAST framework is also useful for refining the quality of clusters.

\section{Experiments}\label{experiment}

We evaluate our unsupervised self-learning method on cross-domain Person Re-ID tasks. Three common large-scale person Re-ID datasets are used, Market-1501~\cite{market1501}, DukeMTMC-Re-ID~\cite{duke}, and CUHK03~\cite{cuhk03}.

\textit{Market-1501}~\cite{market1501} contains 32,668 labelled images of 1,501 identities taken by 6 cameras, which are detected and cropped via Deformable Part Model (DPM)~\cite{DPM}. The dataset is split into training set with 12,936 images of 751 identities and test set with 19,732 images of 750 identities. 

\textit{DukeMTMC-Re-ID}~\cite{duke} consists of 36,411 labelled images belonging to 1,404 identities observed by 8 camera views. As the format of Matket-1501 dataset, it has 16,522 images of 702 identities for the training set and the remaining 19,889 images of 702 identities for the test set. 
Hereafter \emph{Duke} refers to this dataset.

\textit{CUHK03}~\cite{cuhk03} is composed of 14,096 images from 1,467 identities captured by 2 cameras. This dataset was constructed by both manual labelling and DPM. In this work, we experiment on the images detected using DPM. To be in consistency with the protocol of Market-1501 and Duke, new train/test evaluation protocol~\cite{rerank} are used: 7,365 images with 767 identities for training and the remaining 6,732 images with 700 identities for testing.

\subsection{Implementation Details}
\textbf{Model and Preprocessing.} We adopt PCB~\cite{PCB} as our model structure, in which ResNet-50~\cite{resnet50} without last classification layer is used as backbone model. Similar as EANet~\cite{EANet}, we use $9$ regions for feature representation. Instead of using part aligned pooling~\cite{EANet}, we change to use even parts like PCB for simplification. The dimension of each embedding layer is set to $256$. Following each embedding layer, we also implement the classifier layer with one fully connected layer in the promoting stage. The classifier output changes according to the number of clusters generated from HDBSCAN clustering process.

All input images are resized to $384\times128\times3$. It is noting that we only apply random flipping as data augmentation.

\textbf{Training Settings.} We use the SGD optimizer with a momentum of $0.9$ and weight decay of $5\times10^{-4}$ to train the model. Without otherwise specification, in all experiments we set batch size to $64$ and the iteration step to $4$. 
Instead of directly using same learning rates for both conservative and promoting stage, we believe that individually setting the specialized learning rates can work better for our PAST framework. 
The reason is that the parameters from the conservative stage should be updated slower in the promoting stage for avoiding error amplification caused by Softmax cross-entropy loss. Specifically, the learning rate is initialized to $10^{-4}$ on fine-tune layers and $2\times10^{-4}$ on embedding layers in the conservative stage, while for the promoting stage, newly added classifier layers use an initial learning rate of $10^{-3}$ and all other layers $5\times10^{-5}$. 
After $3$ iterations, all learning rates are multiplied by $0.1$. The margin hyper parameter $m$ is set to $0.3$ in both Eq.~\eqref{eq:CTL} and Eq.~\eqref{eq:RTL}.

\textbf{Evaluating Settings.} For performance evaluation, feature vectors from embedding layers of $9$ parts are normalized separately and then concatenated as the output representation. Given a query image, we calculate cosine distance with all gallery images and then sort it as final ranking result. We utilize the Cumulated Matching Characteristics (CMC)~\cite{CMC} and mean Average Precision (mAP)~\cite{market1501} as the performance evaluation measures. CMC curve shows the probability that a query appears in different size of candidate lists. As for mAP, given a single query, the Average Precision (AP) is computed from the area under its precision-recall curve. The mAP is then calculated as the mean value of AP across all queries. Note that single-shot setting is adopted similar to~\cite{PCB} in all experiments. 

\subsection{Ablation Study}
In this subsection, we aim to thoroughly analyse the effectiveness of each components in our PAST framework.

\textbf{Effectiveness of the Conservative Stage.} As shown in Table~\ref{tab:Effectiveness of every part}, we conduct several experiments to verify the effectiveness of the individual components CTL, RTL and the combination of these two triplet loss functions on the task of M$\rightarrow$D and D$\rightarrow$M. 
First, only with CTL, we improve the performance by $18.49\%$ and $12.14\%$ at Rank-1 accuracy compared with the results from $k$-reciprocal encoding method~\cite{rerank} on M$\rightarrow$D and D$\rightarrow$M respectively. 
Second, we observe that containing only our proposed RTL, the Rank-1 accuracy and mAP increase by $21\%$ and $12.64\%$ for M$\rightarrow$D, while $12.91\%$ and $5.69\%$ on D$\rightarrow$M. This obvious improvement shows that both CTL and RTL are useful for increasing model generalization. And CTL obtains slightly lower performance than RTL.
Then, as described in Eq.~\eqref{eq:C}, we combine CTL and RTL together to jointly optimize model in our conservative stage. It is clear  that we achieve better results on both M$\rightarrow$D and D$\rightarrow$M. Especially for D$\rightarrow$M, we gain $2.38\%$ and $4.42\%$ on Rank-1 and mAP comparing to only using CTL, which shows the significant benefit of our RTL. 
Through this conservative stage, we can learn a relative powerful model for target domain. 

\begin{table}
\footnotesize
\setlength{\belowcaptionskip}{-0.2cm}
\setlength{\abovecaptionskip}{-0cm} 
\begin{center}
\setlength{\tabcolsep}{1mm}{
\begin{tabu} to 1\linewidth {l|c|X[c]|X[c]|X[c]|X[c]}
\hline
\multirow{2}{*}{Method} & \multirow{2}{*}{Stage} & \multicolumn{2}{c|}{M$\rightarrow$D} & \multicolumn{2}{c}{D$\rightarrow$M}\\
\cline{3-6}
& & Rank-1   & mAP   & Rank-1 & mAP \\
\hline
\hline
PCB$^*$~\cite{PCB} (DT)		&- & 42.73 & 25.70 & 57.57 &29.01\\
PCB-R$^*$~\cite{rerank} &- & 49.69 &39.38 & 59.74 &41.93 \\
PCB-R-CTL & C & 68.18 & 49.06 & 71.88 & 46.17 \\
PCB-R-RTL & C & 70.69 & 52.02 & 72.65 & 47.62 \\
PCB-R-CTL+RTL & C & 71.63 & 52.05 & 74.26 & 50.59 \\
PCB-R-PAST &C+P & \textbf{72.35} & \textbf{54.26} & \textbf{78.38} & \textbf{54.62} \\
\hline
\end{tabu}}
\end{center}
\caption{The effectiveness of conservative stage and promoting stage in our proposed Self-training with Progressive Augmentation Framework (PAST). D$\rightarrow$M represents that we use Duke~\cite{duke} as source domain and Market-1501~\cite{market1501} as target domain. $*$ denotes that the results are produced by us. \textbf{DT} means Direct Transfer from PCB with 9 regions. \textbf{R} means applying $k$-reciprocal encoding method~\cite{rerank}. \textbf{CTL} represents clustering-based triplet loss~\cite{batchhardtriplet}, while \textbf{RTL} is our proposed ranking-based triplet loss. Our \textbf{PAST} framework consists of conservative stage and promoting stage that are denoted by \textbf{C} and \textbf{P} respectively.}
\label{tab:Effectiveness of every part}
\end{table}

\textbf{Effectiveness of the Promoting Stage.} However, as illustrated in Figure~\ref{fig:figure1}, there is no further gains even with more training iterations when only using triplet-based loss functions. We believe that it is because during conservative stage, the model only 
sees local structure of data distribution brought by triplet samples. Thus, in our PAST framework, we employ softmax cross-entropy loss as the objective function in the promoting stage to train the model with the conservative stage alternately. Refer to Table~\ref{tab:Effectiveness of every part} again, compared with only using conservative stage, our PAST can further improve mAP and Rank-1 by $2.21\%$ and $0.72\%$ on M$\rightarrow$D task, and $4.03\%$ and $4.12\%$ for D$\rightarrow$M. Meanwhile, from Figure~\ref{fig:visual}, the quality of clusters is also improved with our PAST framework. This shows that the promoting stage does play an important role in model generalization. 

Through the above experiments, different components in our PAST have been evaluated and verified. We show that our PAST framework is not only beneficial for improving model generation but also refining clustering quality.
\begin{table}
\footnotesize
\setlength{\belowcaptionskip}{-0.2cm}
\setlength{\abovecaptionskip}{-0cm}
\begin{center}
\setlength{\tabcolsep}{1mm}{
\begin{tabu} to 0.85\linewidth {l|c|X[c]|X[c]|X[c]|X[c]}
\hline
\multirow{2}{*}{Method} & \multirow{2}{*}{Cluster} & \multicolumn{2}{c|}{M$\rightarrow$D} & \multicolumn{2}{c}{D$\rightarrow$M}\\
\cline{3-6}
& & Rank-1   & mAP   & Rank-1 & mAP \\
\hline
\hline
\multicolumn{1}{l|}{\multirow{3}{*}{PCB-R-CTL}} & K & 44.84 &26.93 & 54.39 &29.94  \\
 & D & 53.73 &36.27 & 67.41 &42.42 \\
 & H & 68.18 &49.06 & 71.88 & 46.17 \\
\hline
\multicolumn{1}{l|}{\multirow{3}{*}{PCB-R-CTL+RTL}} & K & 53.99 &34.46 & 56.26 &32.73  \\
 & D &  67.91 &49.08 & 72.54 &48.06 \\
 & H & 71.63 &52.05 & 74.26 &50.59 \\
\hline
\multicolumn{1}{l|}{\multirow{3}{*}{PCB-R-PAST}} & K & 68.94 &49.97 & 75.48 &51.39 \\
 & D & 71.90 &53.07 & 75.62 &51.70 \\
 & H & \textbf{72.35} &\textbf{54.26} & \textbf{78.38} &\textbf{54.62 }\\
\hline
\end{tabu}}
\end{center}
\caption{The comparison of different clustering methods. \textbf{K}, \textbf{D} and \textbf{H} represents K-means, DBSCAN~\cite{DBSCAN} and HDBSCANRank1{HDBSCAN} clustering method respectively.}
\label{tab:The comparison of different clustering methods}
\end{table}

\textbf{Comparison with Different Clustering Methods.} We evaluate three different clustering methods, \ie,  $k$-means, DBSCAN~\cite{DBSCAN} and HDBSCAN~\cite{HDBSCAN} in the conservative stage. The performance of utilizing these clustering methods under different settings are specified in Table~\ref{tab:The comparison of different clustering methods}. For $k$-means, the number of cluster centroids $k$ is set to $702$ and $751$ on target data of Market-1501 and Duke respectively, which is the same as the number of identities of source training data.
It is clear that HDBSCAN performs better than $k$-means and DBSCAN under either only using conservative stage or whole PAST framework. For instance, using HDBSCAN can achieve mAP $54.26\%$ and Rank-1 $72.35\%$ for M$\rightarrow$D task in PAST framework, which are $4.29\%$ and $3.41\%$ higher than using $k$-means, and $1.19\%$ and $0.45\%$ than using DBSCAN. In addition, we also observe that whatever clustering method we use, our PAST framework always outperforms only using conservative stage. This means that on the one hand, HDBSCAN clustering method has more powerful effect in our framework; on the other hand, our PAST framework indeed provides improvement of feature representation on target domain.

\begin{table*}[htbp]
\footnotesize
\setlength{\abovecaptionskip}{-0.25cm}
\setlength{\belowcaptionskip}{-0.25cm}
\begin{center}
\begin{tabu} to .8791\textwidth {l|X[c]|X[c]|X[c]|X[c]|X[c]|X[c]|X[c]|X[c]}
\hline
\multirow{2}{*}{Method} & \multicolumn{2}{c|}{M$\rightarrow$D} & \multicolumn{2}{c|}{D$\rightarrow$M} & \multicolumn{2}{c|}{C$\rightarrow$M} & \multicolumn{2}{c}{C$\rightarrow$D} \\
\cline{2-9}
& Rank-1   & mAP   & Rank-1 & mAP & Rank-1 & mAP & Rank-1 & mAP\\
\hline
\hline
UMDL~\cite{UMDL}'16 & 18.5 & 7.3 & 34.5 &  12.4 & - & - & - & - \\
PUL~\cite{PUL}'18 & 30.0 & 16.4 & 45.5 & 20.5 & 41.9 & 18.0 & 23.0 & 12.0\\
PTGAN~\cite{PTGAN}'18 & 27.4 & - & 38.6 & - & 31.5 & - & 17.6 & - \\
SPGAN~\cite{SPGAN}'18 & 46.4 & 26.2 & 57.7 & 26.7 & - & - & - & - \\
TJ-AIDL~\cite{TJ-AIDL}'18 & 44.3 & 23.0 & 58.2 & 26.5 & - & - & - & - \\
HHL~\cite{HHL}'18 & 46.9 & 27.2 & 62.2 & 31.4 & 56.8 & 29.8 & 42.7 & 23.4 \\
ARN~\cite{ARN}'18 & 60.2 & 33.4 & 70.3 & 39.4 & - & - & - & -\\
EANet~\cite{EANet}'19 & 67.7 & 48.0 & \textcolor{blue}{78.0} & 51.6& \textcolor{blue}{66.4} & \textcolor{blue}{40.6} & \textcolor{blue}{45.0} & \textcolor{blue}{26.4} \\
Theory~\cite{theory}'18 & \textcolor{blue}{68.4} & \textcolor{blue}{49.0} & 75.8 & \textcolor{blue}{53.7} & - & - & - & -\\
\hline
PCB$^*$~\cite{PCB} (DT)'18 & 42.73 & 25.70 & 57.57 & 29.01 & 51.43 & 27.28 &	 29.40 & 16.72\\
PCB-R$^*$~\cite{rerank} & 49.69 & 39.38 & 59.74 & 41.93 & 55.91 & 38.95 & 35.19 & 26.89 \\
\hline
PCB-R-CTL+RTL (Ours) & 71.63 & 52.05 & 74.26 & 50.59 & 77.70 & 54.36 & 65.71 & 46.58  \\
PCB-R-PAST (Ours) & \textcolor{red}{\textbf{72.35}} & \textcolor{red}{\textbf{54.26}} & \textcolor{red}{\textbf{78.38}} & \textcolor{red}{\textbf{54.62}} & \textcolor{red}{\textbf{79.48}} & \textcolor{red}{\textbf{57.34}} & \textcolor{red}{\textbf{69.88}} & \textcolor{red}{\textbf{51.79}}\\
\hline
\end{tabu}
\end{center}
\caption{Comparison with state-of-the-art methods under unsupervised cross-domain setting. In each column, the \textcolor{red}{\textbf{1st}} and \textcolor{blue}{2nd} highest scores are marked by \textcolor{red}{\textbf{red}} and \textcolor{blue}{blue} respectively. D, M, C represent Duke~\cite{duke}, Market-1501~\cite{market1501} and CUHK03~\cite{cuhk03} respectively.}
\label{tab:cross_domain_sota}
\end{table*}

\subsection{Comparison with State-of-the-art Methods}
Following evaluation setting in~\cite{EANet, HHL}, we compare our proposed PAST framework with state-of-the-art unsupervised cross-domain methods, shown in Table~\ref{tab:cross_domain_sota}.
It can be seen that only using conservative stage with CTL and RTL for training, the performance is already competitive with other cross-domain adaptive methods. 
For example, although EANet~\cite{EANet} proposes complex part-aligned pooling and combines pose segmentation to provide more information for adaptation, our conservative stage still outperforms it by $3.93\%$ in Rank-1 and $4.05\%$ in mAP when testing on M$\rightarrow$D. 
Moreover, our PAST framework surpasses all previous methods by a large margin, which achieves $54.26\%$, $54.62\%$, $57.34\%$, $51.79\%$ in mAP and $72.35\%$, $78.38\%$, $79.48\%$, $69.88\%$ in Rank-1 accuracy for M$\rightarrow$D, M$\rightarrow$D, C$\rightarrow$M, C$\rightarrow$D. 
We can also prove that it is useful to alternately use conservative and promoting stage by comparing with the last two rows in Table~\ref{tab:cross_domain_sota}. 
Especially, our PAST can improve $4.71\%$ and $5.21\%$ in Rank-1 and mAP for C$\rightarrow$D compared with only using conservative stage.

\begin{figure*}[h]
\centering
\includegraphics[trim =0mm 0mm 0mm 0mm, clip, width=1\linewidth]{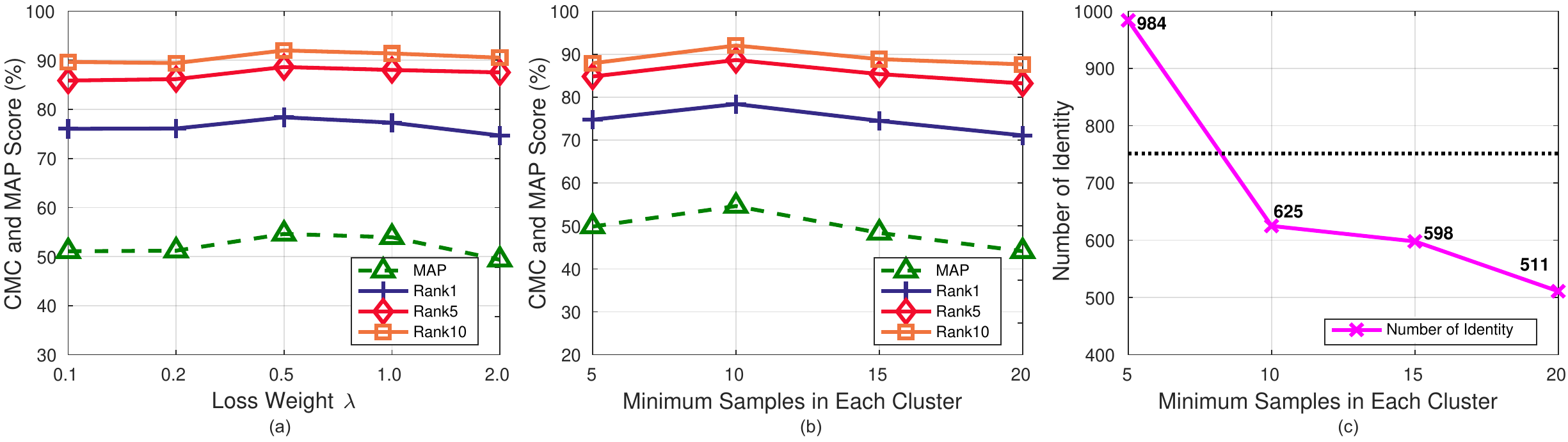}
\setlength{\abovecaptionskip}{-0.35cm} 
\setlength{\belowcaptionskip}{-0.6cm}
\caption{Analysis of hyper parameters on D$\rightarrow$M setting. (a): The impact of the loss weight $\lambda$; (b): The impact of the minimum samples $S_{\min}$ at each cluster in HDBSCAN clustering method; (c): The number of clusters from HDBSCAN with different minimum sample $S_{\min}$.}
\label{fig:lw_ms}
\end{figure*}

\subsection{Parameter Analysis}
Besides, we conduct additional experiments to evaluate the parameter sensitivity.

\textbf{Analysis of the Loss Weight $\lambda$.}
$\lambda$ is a hyper parameter which is used to trade off the effect between ranking-based triplet loss (RTL) and clustering-based triplet loss (CTL). 
We evaluate the impact of $\lambda$, which is sampled from $\lbrace 0.1, 0.2, 0.5, 1.0, 2.0 \rbrace$, on the task of D$\rightarrow$M. 
The results are shown in Figure~\ref{fig:lw_ms} (a). 
We observe that the best result is obtained when $\lambda$ is set to $0.5$. Note that large or small $\lambda$ has limitation on the improvement of performance.

\textbf{Analysis of the Minimum Samples $S_{\min}$.}
In addition, we analyse how the number of minimum samples ($S_{\min}$) for every cluster in HDBSCAN clustering affects the Re-ID results. 
We test the impact of $\lbrace 5, 10, 15, 20\rbrace$ minimum samples on the performance of our PAST framework on D$\rightarrow$M setting. 
As shown in Figure~\ref{fig:lw_ms} (b), we can see that setting $S_{min}$ to $10$ yields superior accuracy. 
Meanwhile, different $S_{\min}$ has large variance on the final number of pseudo identities from HDBSCAN. We believe that it is because samples from the same class will be separated to several clusters when $S_{\min}$ is too small, while low-density classes will be abandoned if $S_{\min}$ is too large. 
This can be verified from Figure~\ref{fig:lw_ms} (c), the number of identity from HDBSCAN with minimum sample $10$ is \textbf{$625$}, which is the closest one to the true value $751$ in Market-1501 training set.
\section{Conclusion}
In this paper, we have presented a self-training with progressive augmentation framework (PAST) for unsupervised cross-domain person re-identification. Our PAST consists of two different stages, \ie, the conservative and promoting stage, which are adopted alternately to offer complementary information for each other. 
Specifically, the conservative stage mainly captures local information with triplet-based loss functions, while the promoting stage is used for extracting global information. 
For alleviating the dependence on clustering quality, we also propose a novel label-free ranking-based triplet loss. 
With these proposed method, the model generalization gains significant improvement, as well as the capability of feature representation on target domain. 
Extensive experiments show that our PAST outperforms the state-of-the-art unsupervised cross-domain algorithms by a large margin.

We plan to extend our work to other unsupervised cross-domain applications, such as face recognition 
and image retrieval tasks.
\textbf{Acknowledgements}
This work was in part supported by the Natural Science Foundation of Shanghai, China under Grand \#17ZR1431500.

{\small
\bibliographystyle{ieee}
\bibliography{egbib}
}

\section*{Appendix}
\setcounter{section}{0} 
\section{More Experimental Results}
\subsection{More Experiments for Parameter Analysis}
\textbf{Analysis of the Maximum Ranking Position $\eta$ for Positive Sample.}
The maximum ranking position $\eta$ is a tunable hyper-parameter in the ranking-based triplet loss (RTL), as shown in Eq.~$\left(4\right)$ in the main paper, which defines the range $(0, \eta]$ for selecting positive samples and the range $(\eta, 2\eta]$ for negative samples. 
We conduct several experiments to evaluate the sensitivity of our method to $\eta$ when transferring from Duke~\cite{duke} to Market-1501~\cite{market1501}, as shown in Table~\ref{fig:position}. 
It shows that when $\eta$ is equal or larger than $20$, we can obtain nearly same and competitive results. 
And we set $\eta=20$ in all experiments except this part. The performance drops quickly when $\eta$ is extremely small. 
We believe that it is due to the unbalanced identities, \eg, the minimal and maximal numbers of images are $2$ and $72$ respectively in the training set of Market-1501, which results in a large probability that the selected positive and negative samples are from the same (ground-truth) identity.

\begin{table}[b]
\setlength{\belowcaptionskip}{-0.1cm}
\setlength{\abovecaptionskip}{-0.2cm} 
\begin{center}
\begin{tabular}{ccccc}
\hline
\multicolumn{5}{c}{D$\rightarrow$M}                                         \\ \hline \hline
\multicolumn{1}{c|}{$\eta$} & \multicolumn{1}{c|}{Rank-1} & \multicolumn{1}{c|}{Rank-5} & \multicolumn{1}{c|}{Rank-10} & \multicolumn{1}{c}{mAP}   \\ \hline
\multicolumn{1}{c|}{5}  & \multicolumn{1}{c|}{71.85}  & \multicolumn{1}{c|}{83.22}  & \multicolumn{1}{c|}{87.00}   & 44.37 \\
\multicolumn{1}{c|}{10} & \multicolumn{1}{c|}{73.78}  & \multicolumn{1}{c|}{84.09}  & \multicolumn{1}{c|}{87.62}   & 47.64 \\
\multicolumn{1}{c|}{15} & \multicolumn{1}{c|}{77.43} & \multicolumn{1}{c|}{86.70}  & \multicolumn{1}{c|}{89.99}   & 51.77 \\
\multicolumn{1}{c|}{20} & \multicolumn{1}{c|}{\textbf{78.38}}  & \multicolumn{1}{c|}{\textbf{88.63}}  & \multicolumn{1}{c|}{\textbf{92.01}}   & \textbf{54.62}  \\
\multicolumn{1}{c|}{25} & \multicolumn{1}{c|}{78.27} & \multicolumn{1}{c|}{88.63}  & \multicolumn{1}{c|}{91.63}   & 55.27 \\
\multicolumn{1}{c|}{30} & \multicolumn{1}{c|}{78.15}  & \multicolumn{1}{c|}{88.93}  & \multicolumn{1}{c|}{91.95}   & 54.46 \\
\multicolumn{1}{c|}{35} & \multicolumn{1}{c|}{78.59}  & \multicolumn{1}{c|}{88.48}  & \multicolumn{1}{c|}{91.83}   & 55.10 \\ \hline
\end{tabular}
\end{center}
\caption{The influence of maximum ranking position $\eta$ for triplet selection of RTL in our PAST framework on D$\rightarrow$M setting.}
\label{fig:position}
\end{table}

\subsection{More Qualitative Analyses}
\begin{figure*}[tbp]
\centering
\includegraphics[trim =0mm 0mm 0mm 0mm, clip, width=1\linewidth]{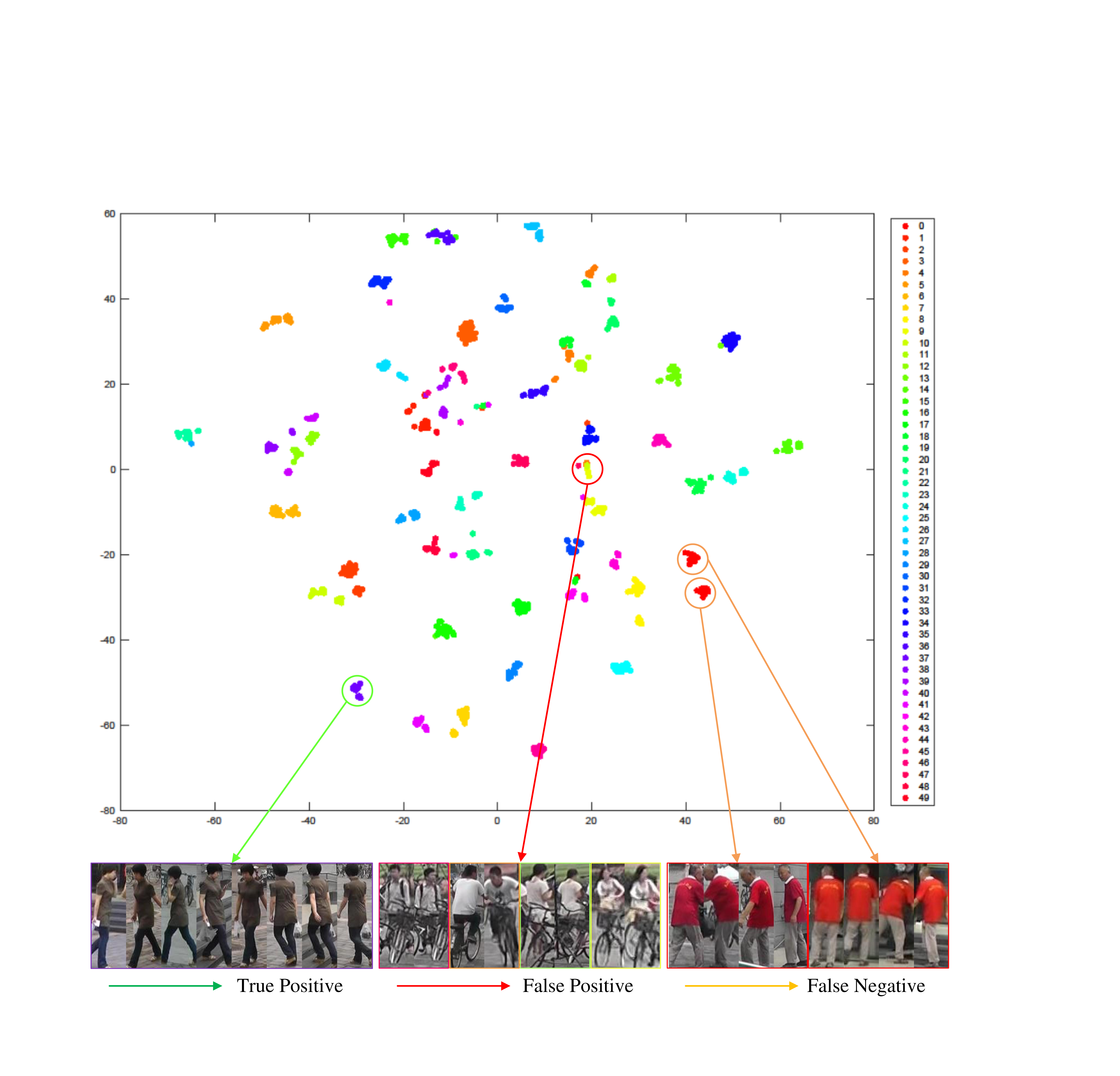}
\caption{Qualitative analysis of the feature representation by using t-SNE~\cite{tsne} visualization on a subset of Market-1501~\cite{market1501} training data. According to the clustering result, we choose the Top-50 identities which contain Top-50 the largest number of images. Points with the same color have the same (ground-truth) identity. The \textcolor{green}{\textbf{green circle}} means images from the same identity are gathered together, and the cluster is extremely reliable. Images in \textcolor{orange}{\textbf{orange circle}} are both from same identity, yet they are clustered to two different classes. We can see that due to the camera style, images from the two classes have different appearances. In the \textcolor{red}{\textbf{red circle}}, although our algorithm may gather images from different (ground-truth) identities into the same cluster, these images usually share very similar appearances and are hard to distinguish with each other. For instances, every image in the red circle contains one person with white clothes and a black bicycle.}
\label{fig:tsne}
\end{figure*}

\textbf{Qualitative Analysis of the Feature Representation.}
To demonstrate the results intuitively, we visualize the feature embeddings calculated by our PAST framework in 2-D using t-SNE~\cite{tsne}. Three representative classes are displayed by showing the corresponding images in the bottom, \ie, \textit{true positive samples}, \textit{false positive samples} and \textit{false negative samples}. As illustrated in Figure~\ref{fig:tsne}, images belonging to the same identity are almost well gathered together, while those from different classes usually stay apart from each other. It implies that our PAST framework can improve the capability of model generalization which is beneficial for learning discriminative feature representation on the target-domain dataset.

\begin{figure*}[tbp]
\centering
\includegraphics[trim =0mm 0mm 0mm 0mm, clip, width=1\linewidth]{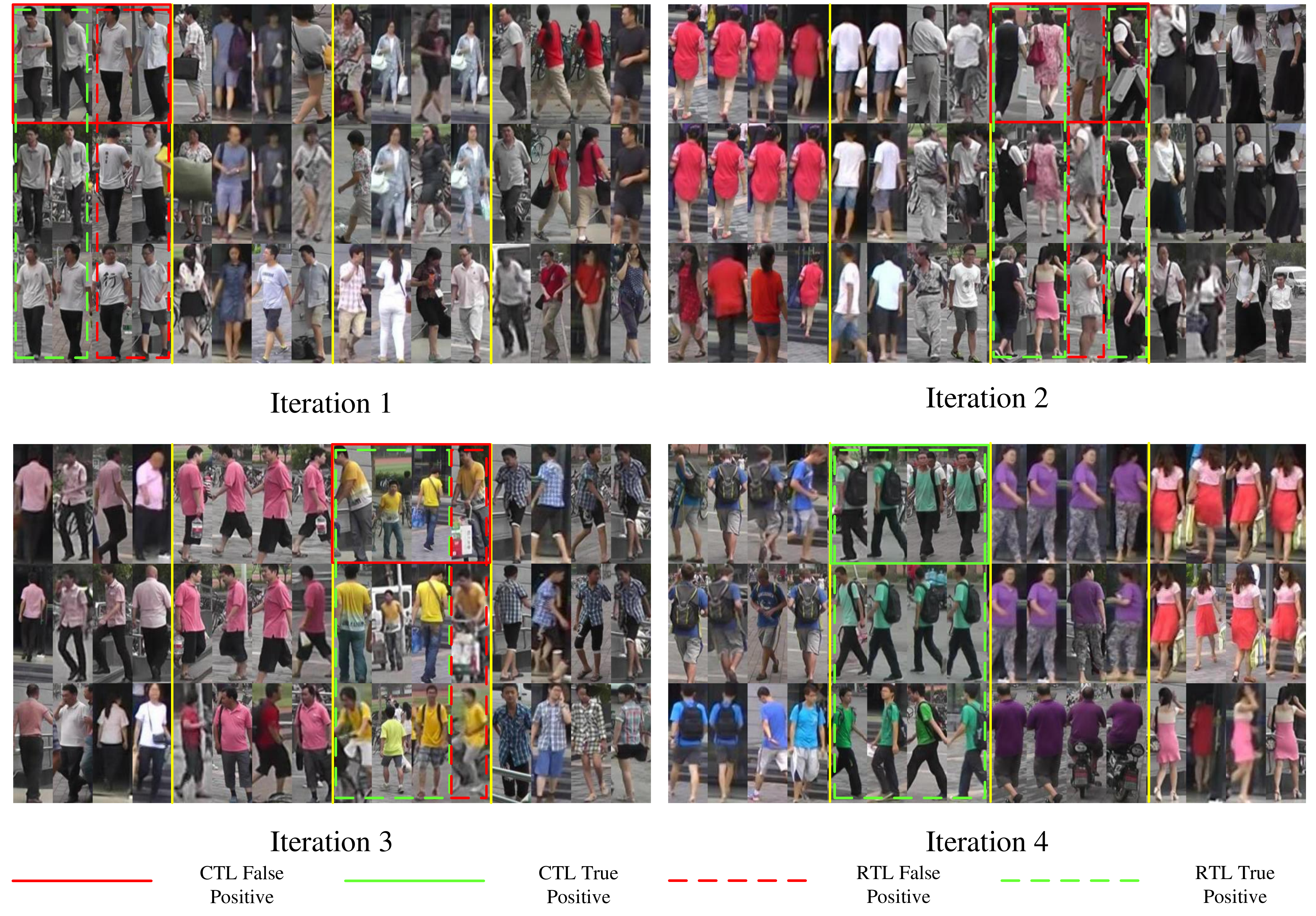}
\caption{Quality of the triplet selection over training iterations. Images from different clusters are divided by \textcolor{yellow}{\textbf{yellow line}}. The \textcolor{red}{\textbf{red line}} means generated triplets are not completely correct, while \textcolor{green}{\textbf{green line}} represents generated triplets are completely correct. 
The solid line and dashed line are for triplets, which are generated from CTL and RTL respectively. 
We use Duke~\cite{duke} as the source domain and Market-1501~\cite{market1501} as the target domain.}
\label{fig:triplets}
\end{figure*}

\textbf{Qualitative Analysis of the Triplet Selection.}
In Figure~\ref{fig:triplets}, we visualize the triplet samples generated in the conservative stage for CTL and RTL, respectively. We summarize the main advantages of the proposed PAST method in the following.
\begin{enumerate}
\item The proposed PAST algorithm can significantly improve the quality of the clustering assignments during training.
As shown in the first row of the iterations from 1 to 4, the images assigned to the same class by the proposed method tend to be more and more similar. On the other hand, the quality of the pseudo labels assigned to each images is steadily improved during training. 
It means that our PAST framework is beneficial for learning discriminating
feature presentation and can assign more reliable pseudo labels to target images. 
The accurate pseudo labels can be used to promoting stage to improve the model generalization further.
\item RTL is useful for remedying the variance caused by CTL. Refer to Figure~\ref{fig:triplets} again, we can observe that the third cluster in iteration 2 is noisy and the selected triplets from CTL are not faithful. 
However, RTL can select correct positive sample even the cluster is dirty. 
We believe that the reason is that RTL just depends on the similarity ranking matrix and the top $\eta$ similar images are used for generating positive samples, which is more reliable when the features representation is not so discriminative. 
\item RTL helps to further optimize the network, especially in the later iteration. From Figure~\ref{fig:triplets}, we can also see that different clusters in one mini-batch may look different due to unique color of clothes, which results in extremely simple negative samples and slows down the optimization when training on CTL. 
Whereas, considering the triplets generated from the RTL, negative images are extremely similar to the anchors, which is even hard to be well recognized by human beings. For example, at the second column in iteration 4, all images look like one person, although images from the first two rows are same person, while those from the third row belong to another person. 
\end{enumerate}

\end{document}